\documentclass[sn-mathphys,Numbered]{sn-jnl}

\usepackage{graphicx}%
\usepackage{multirow}%
\usepackage{amsmath,amssymb,amsfonts}%
\usepackage{amsthm}%
\usepackage{mathrsfs}%
\usepackage[title]{appendix}%
\usepackage{xcolor}%
\usepackage{textcomp}%
\usepackage{manyfoot}%
\usepackage{booktabs}%
\usepackage{algorithm}%
\usepackage{algorithmicx}%
\usepackage{algpseudocode}%
\usepackage{listings}%

\theoremstyle{thmstyleone}%
\theoremstyle{thmstyletwo}%

\theoremstyle{thmstylethree}%

\begin{document}

\title[Article Title]{AaKOS: Aspect-adaptive Knowledge-based Opinion Summarization}


\author[1]{\fnm{Guan} \sur{Wang}}\email{guan.wang@autuni.ac.nz}

\author*[1]{\fnm{Weihua} \sur{Li}}\email{weihua.li@aut.ac.nz}

\author[1]{\fnm{Edmund M-K.} \sur{Lai}}\email{edmund.lai@aut.ac.nz}

\author[2]{\fnm{Quan} \sur{Bai}}\email{quan.bai@utas.edu.au}

\affil*[1]{\orgdiv{School of Engineering, Computer \& Mathematical Sciences}, \orgname{Auckland University of Technology}, \orgaddress{\city{Auckland}, \postcode{1010}, \country{New Zealand}}}

\affil[2]{\orgdiv{School of Information and Communication Technology}, \orgname{University of Tasmania}, \orgaddress{\street{Churchill Ave}, \city{Hobart}, \postcode{7005}, \state{Tasmania}, \country{Australia}}}

\abstract{

The rapid growth of information on the Internet has led to an overwhelming amount of opinions and comments on various activities, products, and services. This makes it difficult and time-consuming for users to process all the available information when making decisions. Text summarization, a Natural Language Processing (NLP) task, has been widely explored to help users quickly retrieve relevant information by generating short and salient content from long or multiple documents. Recent advances in pre-trained language models, such as ChatGPT, have demonstrated the potential of Large Language Models (LLMs) in text generation. However, LLMs require massive amounts of data and resources and are challenging to implement as offline applications. Furthermore, existing text summarization approaches often lack the ``adaptive" nature required to capture diverse aspects in opinion summarization, which is particularly detrimental to users with specific requirements or preferences. In this paper, we propose an Aspect-adaptive Knowledge-based Opinion Summarization model for product reviews, which effectively captures the adaptive nature required for opinion summarization. The model generates aspect-oriented summaries given a set of reviews for a particular product, efficiently providing users with useful information on specific aspects they are interested in, ensuring the generated summaries are more personalized and informative. Extensive experiments have been conducted using real-world datasets to evaluate the proposed model. The results demonstrate that our model outperforms state-of-the-art approaches and is adaptive and efficient in generating summaries that focus on particular aspects, enabling users to make well-informed decisions and catering to their diverse interests and preferences. 
}

\keywords{Aspect-adaptive Opinion Summarization, Text Summarization, Knowledge Graph, Deep Learning}

\maketitle

\section{Introduction}\label{sec1}

The rapid development of the Internet generates a massive amount of information daily, leading to challenges in efficiently retrieving useful information, such as online shopping reviews. Reviews not only provide important information to users for making informed decisions but also enable enterprises to adjust their marketing strategies. As for individuals, they may focus on different aspects of a product or service \cite{amplayo-etal-2021-aspect}.

Text summarization techniques can condense salient information from multiple comments, enabling users to make efficient decisions \cite{amplayo-etal-2021-aspect}. The objective of text summarization is to generate concise summaries while preserving core information. There are two main methods that are commonly used in the field of text summarization: extractive and abstractive methods. \textbf{Extractive} methods identify the most meaningful phrases and sentences and combine them without modification to form a text summary \cite{zhong2020extractive,xu2019discourse}. In contrast, \textbf{abstractive} methods generate summaries with novel language, free from constraints \cite{dou2020gsum,see2017get}. Hybrid approaches have also been proposed, combining extractive and abstractive methods to generate coherent text while retaining key information \cite{zhang2020pegasus,liu2019text}. Opinion summarization is acknowledged as a sub-research domain of text summarization, which aims to generate summaries from a set of reviews \cite{lu2009rated,amplayo-etal-2021-aspect,ahuja-etal-2022-aspectnews}. Opinion summarization has been studied using extractive and abstractive approaches, where ranking algorithms, rule-based methods, and machine learning are employed to identify important phrases, sentences, or paragraphs \cite{raut2014opinion,condori2017opinion}. Abstractive approaches utilize neural networks and deep learning technologies to generate more coherent summaries \cite{amplayo-etal-2021-aspect,angelidis-lapata-2018-summarizing}.

Recent advances in text summarization have introduced self-supervised methods that employ similar content as pseudo summaries in lieu of gold-standard summaries \cite{bravzinskas2019unsupervised,im-etal-2021-self,amplayo-etal-2021-aspect}. Although these methods facilitate the training of summarization models without the need for labor-intensive and expensive human-generated summaries, they lack the ``adaptive" nature that is essential for capturing diverse aspects in opinion summarization. One major concern with using similar content as pseudo summaries is the potential loss of salient information, which is particularly detrimental to users with specific requirements or preferences. This limitation leads to incomplete or inadequate summaries, as not all critical aspects may be represented. Consequently, users may not have access to all the necessary information for making informed decisions. Moreover, when using sampled reviews as pseudo summaries, there is a risk that the generated summaries may not comprehensively cover the diverse aspects present in a larger set of reviews. The lack of adaptability is particularly problematic for users interested in specific aspects or features. This underscores the importance of developing summarization methods that can effectively capture and present a wide range of aspects to cater to users' diverse needs and interests.

To address the aforementioned limitations, in this paper, we propose the Aspect-adaptive Knowledge-based Opinion Summarization (AaKOS) model, which can effectively capture the adaptive nature required for opinion summarization. By incorporating adaptability in AaKOS, it can dynamically adjust to users' preferences and interests, ensuring the generated summaries are more personalized and informative. On top of that, AaKOS does not rely on datasets with text-golden summary pairs, which are difficult and expensive to create. Instead, AaKOS works with datasets without human-written summaries as labels, allowing for greater datasets flexibility. Specifically, AaKOS transforms plain texts into weighted Knowledge Graphs and encodes both aspects and corresponding graphs using a text encoder and a graph encoder, respectively. The text encoder generates embeddings of aspects and extracts aspects to form the Knowledge Graphs. AaKOS model can be trained using a self-supervised approach, which involves utilizing the original sentences that the aspects and knowledge graph are derived from as pseudo summaries. In contrast to other self-supervised models that rely on synthetic pairs of source texts and pseudo summaries, AaKOS does not sample reviews from the dataset. Instead, it accurately pairs knowledge graphs and aspects with their corresponding content.

AaKOS significantly enhances output precision, particularly for given aspects. When an aspect is determined, related content is extracted from the knowledge graph, leading to the formation of a sub-graph. Consequently, the output contains information only about the specific aspect from the graph, ensuring that other aspects are not summarized in the output. This adaptability enables the AaKOS model to generate aspect-specific summaries, catering to users with diverse interests and preferences, and providing more personalized and informative results.

\begin{figure}
\centering
	\includegraphics[width=0.8\textwidth]{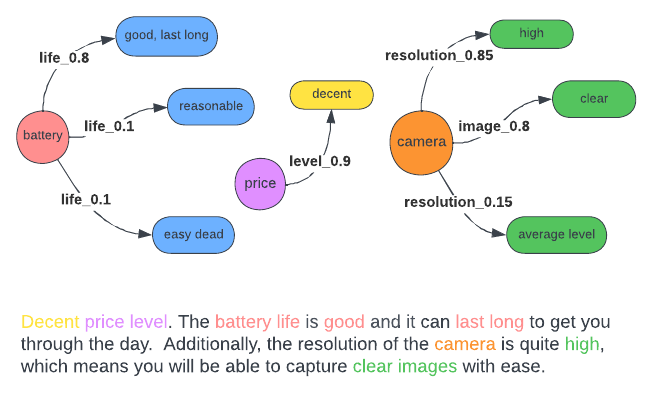}
	\caption{With particular aspects, our model can generate a corresponding summary for a mobile phone from the dataset. Aspects and related contents are coloured in the summary. The value of the weight controller is set to $wc>0.2$.} \label{fig1}
\end{figure}

Figure~\ref{fig1} demonstrates the summary generated by our model when given several specific aspects. By leveraging knowledge-based techniques and aspect-adaptive summarization, our proposed model addresses the limitations of existing methods and offers a more efficient way to extract and present relevant information to users. This approach not only helps users make well-informed decisions but also supports enterprises in refining their strategies based on the aspects that matter most to their customers.

The rest of this paper is organized as follows: Section 2 provides an overview of related work and discusses the limitations of existing approaches. In Section 3, we present our proposed novel opinion summarization model in detail. Sections 4 and 5 describe the experimental setup and discuss the results obtained from the experiments, respectively. Lastly, Section 6 offers the conclusions and future works of the work presented in this paper. 

\section{Related works}\label{sec2}

\subsection{Text Summarization}

Numerous research efforts have focused on enhancing abstractive generators' paraphrasing capabilities. For example, Rush et al. apply the neural encoder-decoder architecture to text summarization and discussed potential encoder choices \cite{rush2015neural}. Based on this research work, many researchers investigated approaches to improve encoding and decoding capabilities, addressing issues such as out-of-vocabulary and repetition. See et al., for example, employ pointer and coverage mechanisms to tackle these problems while developing the pointer-generator network to accurately reproduce source text information \cite{see2017get}. Gehrmann et al. propose a two-step process to address content selection issues: token-level sequence tagging for content selection and bottom-up copy attention to restrict attention over selected source text fragments \cite{gehrmann2018bottom}.

In recent years, fact-aware summarization has attracted significant attention, highlighting the challenges associated with generating factually accurate summaries. Previous studies have revealed that abstractive summarization models are prone to hallucinating phenomena, with approximately 30\% of summaries from state-of-the-art models exhibiting factual inconsistency \cite{kryscinski2019evaluating, cao2018faithful}. To mitigate this issue, various approaches have been proposed. Cao et al. introduce fact-aware neural abstractive summarization, incorporating extracted facts into the encoder alongside the source text \cite{cao2018faithful}. Kryscinski et al. develop a novel method for verifying factual consistency and identifying conflicts between the source text and summaries \cite{kryscinski2019evaluating}. Li et al. treat fact-aware summarization as an entailment-aware process, arguing that summaries should be semantically entailed by the source text \cite{li2018ensure}. Zhu et al. utilize Knowledge Graphs to integrate factual information into the summarization process \cite{zhu-etal-2021-enhancing}. 

Despite the advancements in abstractive summarization and fact-aware techniques, existing works possess certain limitations that stem from overlooking the adaptive nature of opinion summarization. The ability to adapt summaries based on user-specific preferences, requirements, or interests is essential for capturing a wide range of aspects and accommodating diverse user needs. Furthermore, most existing approaches predominantly focus on generating summaries that preserve the general or most salient information from the source text, often leading to summaries that may not cater to individual user preferences. 

In our preliminary research work, the Knowledge-aware Abstractive Text Summarization (KATSum) model was proposed to address the issue of noise information in datasets by transforming text data into Knowledge Graphs \cite{wang2022katsum}. The Knowledge Graphs are employed to filter out the noisy and identify useful triplets, which then guide the text summarization generation process. Despite its effectiveness in dealing with noise, this model has a notable limitation: it is supervised and therefore unsuitable for datasets that lack golden summaries. Obtaining these golden summaries can be a costly and time-consuming process. Moreover, same as the existing works, the KATSum model does not take into account the adaptive nature required for opinion summarization.

\subsection{Opinion Summarization}
Similar to general text summarization, opinion summarization can be categorized into abstractive and extractive methods. Mirroring the developmental trajectory of general summarization, extractive methods initially gained widespread use in this domain due to the high cost of creating golden summaries for datasets, particularly for review datasets where such summaries are not mandatory \cite{chu2019meansum}. Early works in this field predominantly treated the task as a sentence or phrase selection problem, employing either ranking or classification approaches. For instance, Wei et al. prioritize sentences that closely corresponded to the query \cite{wei2008query}, while Erkan et al. employed a stochastic graph-based method to rank sentences by calculating their importance based on eigenvector centrality within a graph representation of the sentences \cite{erkan2004lexrank}. Presently, extractive methods remain popular for opinion summarization \cite{condori2017opinion,lavanya2020aspect,gamzu2021identifying,amplayo-etal-2021-aspect}.

Abstractive methods for opinion summarization began to emerge around 2010, with Ganesan et al. introducing a graph-based algorithm for generating abstractive summaries \cite{ganesan2010opinosis}. Although this approach produced abstractive summaries, it selected words, phrases, or sentences from the original text, rendering it more akin to an extractive method. In recent years, a growing number of studies have begun to leverage machine learning and deep learning techniques for this task. Notably, Chu et al. presented an unsupervised neural model for multi-document summarization, proposing an end-to-end architecture featuring an auto-encoder. This approach decoded the mean of the input reviews' representations into a coherent summary review without relying on any review-specific attributes \cite{chu2019meansum}.

\subsection{Self-supervision}
In addition to the abstractive and extractive categories, opinion summarization can also be classified into supervised and unsupervised categories. Given the scarcity of golden summaries for datasets, unsupervised approaches are more frequently employed to address this challenge. Both \cite{bravzinskas2019unsupervised} and \cite{chu2019meansum} are unsupervised models that employ auto-encoders, suggesting that review representations can encapsulate sentiment, topics, and opinions about products.

Recently, text summarization has shifted toward a self-supervised approach by utilizing similar content as pseudo summaries \cite{amplayo-etal-2021-aspect,elsahar2021self}. To achieve this, Amplayo et al. fine-tune a pre-trained model using their synthetic training dataset of (review, summary) pairs and generate aspect-specific summaries by modifying their introduced aspect controllers \cite{amplayo-etal-2021-aspect}. Meanwhile,  Elsaha et al. tackle multi-document opinion summarization by assuming one of the documents serves as a target summary for a set of similar documents \cite{elsahar2021self}.

However, sampling similar content or documents from the entire dataset may result in the loss of salient information, as a subset may not encompass all relevant details. In the context of opinion summarization, this lost information could be significant to different individuals with distinct requirements. The sampled reviews may not cover all aspects of a large review set. For example, Amplayo et al. develop a method for generating aspect-specific summaries and constructing a synthetic dataset composed of (review, summary) pairs. To achieve this, they employ a technique involving the sampling of reviews as pseudo summaries, introducing three distinct aspect controllers at the word, sentence, and document levels \cite{amplayo-etal-2021-aspect}. Nevertheless, this approach unavoidably leads to a certain degree of information loss. 


Differing from the existing works, our approach first transforms the text dataset into Knowledge Graphs, which consist of numerous triplets with weighted edges. The edges are formatted as ${attribute\_weight}$, as depicted in Figure~\ref{fig1}. To train our model, we utilize a self-supervised methodology, creating pseudo summaries for the relevant triplet(s) and mapping them as sample pairs. Aspect control is achieved through the input aspect set and the filtered Knowledge Graph, which is based on the aspect set. Additionally, we introduce a Weight Controller $\alpha$ to guide the selection of aspect attributes, enabling the regulation of sentiment trends within these aspects. Consequently, our model can generate opinion summaries tailored to meet diverse requirements related to aspects and sentiment trends.

\section{Aspect-adaptive Knowledge-based Opinion Summarization Model}\label{sec2}

In this section, we present an in-depth description of the proposed Aspect-adaptive Knowledge-based Opinion Summarization (AaKOS) model for tackling the aspect-adaptive opinion summarization task. This task is characterized as a text generation process that utilizes inputs in graph format. These inputs are subsequently transformed into graph embeddings via a graph encoder. In AaKOS, Graph Attention Networks (GATs) \cite{velickovic2018graph} are employed as the means to encode these graphs. This is because GATs lie in the ability to effectively capture the complex relationships between nodes in a graph through attention mechanisms by casting the weighted edges into weighted nodes representing the relationships of two connected nodes. In this way, GATs can adaptively focus on more relevant and informative connections while encoding the graph structure.

Data pre-processing is a key step of AaKOS. Given a collection of reviews related to a single product, our objective is to convert the plain text data into Knowledge Graphs and subsequently utilize these relevant graphs for generating summaries. Prior to implementing the proposed model, it is essential to pre-process the dataset through a series of steps, including cleaning the dataset, pre-training a BERT model, extracting noun chunks, determining the appropriate cluster numbers, clustering all reviews based on sentence embeddings derived from the pre-trained BERT model, identifying aspects, and ultimately extracting pertinent triplets to construct Knowledge Graphs. 

\begin{figure}
\centering
    \includegraphics[width=0.8\textwidth]{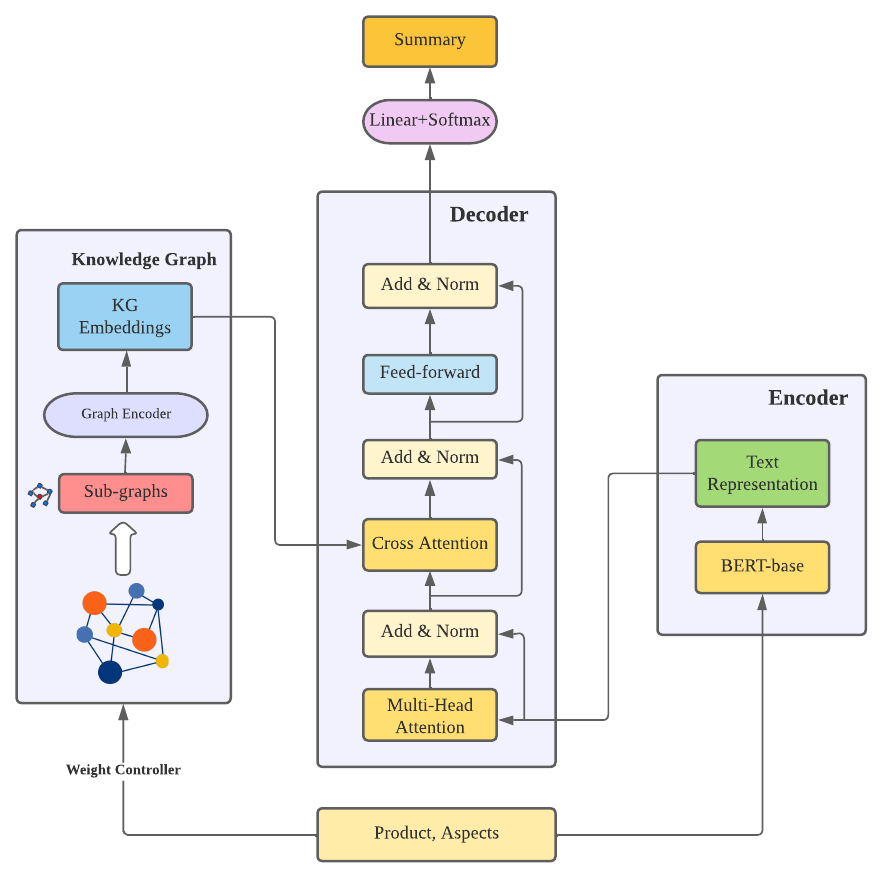}
    \caption{The architecture of the model. A weight controller is introduced while retrieving relative parts of graphs. If the edge weight of a triplet is bigger than the controller, this triplet will be selected and its corresponding text records will be selected as a pseudo summary.} \label{fig2}
\end{figure}

The architecture of AaKOS is illustrated in Figure~\ref{fig2}, which comprises two encoders and a decoder. The two encoders consist of a BERT-based text encoder \cite{devlin2018bert} and a graph encoder. The former utilizes a pre-trained BERT model \cite{devlin2018bert} to generate hidden states from a given set of aspects, which are then employed as part of the decoder input. The latter transforms the filtered sub-graphs into graph embeddings. The decoder is constructed with multi-head attention layers for text and cross-attention layers for integrating text embeddings and graph embeddings. The cross-attention mechanism can be represented by the following equations:

\begin{equation}\label{eq:cross attention 1}
		Q = W_qE^{g}, K=W_kE^{t}, V=W_vE^{t},
\end{equation}

 \begin{equation}\label{eq:cross attention 2}
		Attn=softmax(\frac{QK^{T}}{\sqrt{d_k}})V,
\end{equation}

\noindent

\noindent
where $W_q, W_k, W_v$ represent learnable parameters. $E^{g}$ and $E^{t}$ denote the embeddings of graph and text, respectively. Additionally, $d_k$ corresponds to the embedding dimension. In AaKOS model, Graph Attention Networks (GATs) \cite{velickovic2018graph} are employed along with an additional global node featuring the product name as the node value. This is mainly due to the outstanding performance in the current setting. Thus, GATs are utilized as the graph encoder in the final design. The attention mechanisms in GATs are computed using Equations \ref{eq:normalized attention}, \ref{eq:attention}, and \ref{eq:attention weight}.

    \begin{equation}\label{eq:normalized attention}
		\alpha_{ij} = softmax_j(\beta_{ij}) = \dfrac{exp(\beta_{ij})}{\sum_{k \in \Gamma(n_i)} exp(\beta_{ik})},
	\end{equation}	
	
	\begin{equation}\label{eq:attention}
		\beta_{ij} = LeakyReLU(a^T[Wh_i||Wh_j]),
	\end{equation}	
	
	\begin{equation}\label{eq:attention weight}
		h_i = \sigma(\sum_{j \in N_i}\alpha_{ij}Wh_j),
	\end{equation}	
	
\noindent where $\Gamma(n_i)$ denotes the neighbourhood of node $n_i$ within the graph, while $h_i$ corresponds to the representations of $n_i$. $\beta_{ij}$ signifies the importance of $n_j$ with respect to $n_i$, and $\alpha_{ij}$ represents the normalized attention of $n_i$ across all neighbouring nodes. The final graph embeddings, denoted as $\bf{emb} \in R^{d}$, are gathered by the additional global node, which connects every node in the graph.

In this model, all reviews are initially converted into knowledge graphs, which are subsequently mapped into a lower-dimension vector space using GATs. The resulting vectors (graph embeddings) are utilized in conjunction with aspect embeddings from a text encoder as input for a decoder to generate summaries.

To regulate the sentiment trend, the Knowledge Graph is designed as a weighted structure, and a parameter called the Weight Controller is introduced. This parameter serves as a threshold for filtering relevant attributes by comparing edge weights to the controller's value. Adjusting the weight controller enables the filtering of less-discussed perspectives, thereby highlighting aspects that most people focus on or identifying prevailing trends. For example, in Figure~\ref{fig1}, the weight controller is set to $wc > 0.2$, excluding attributes with a weight lower than 0.2 from the summary.

\section{Experimental Setup}\label{sec3}

\subsection{Data set}\label{subsec1}
    
To conduct the experiments, three real-world datasets are employed to evaluate the performance of the proposed AaKOS model, i.e., Amazon Product Review dataset \footnote{https://s3.amazonaws.com/amazon-reviews-pds/tsv/index.txt}, SPACE dataset \cite{angelidis2021extractive} and YELP dataset \footnote{https://www.yelp.com/dataset}. 

\begin{itemize}
    \item \textbf{Amazon Product Review} dataset is an extensive collection of millions of product reviews from Amazon, encompassing 46 categories. We focus on the top 10 categories with the highest number of reviews. Each review in the dataset includes 15 attributes, such as marketplace, customer\_id, star\_rating, review\_id, and review\_date, among others. In this study, we retain only the essential features, specifically product\_id, review\_headline, and review\_body.

    \item \textbf{SPACE} (Summaries of Popular and Aspect-specific Customer Experiences) dataset is a comprehensive compilation of ``hotel" reviews sourced from TripAdvisor \footnote{https://www.tripadvisor.com/}. It features human-written abstractive opinion summaries intended solely for evaluation purposes.

    \item \textbf{YELP} is a dataset composed of businesses, reviews and user data from Yelp. Chu et al. also collect 200 reference summaries from Amazon Mechanical Turk \footnote{https://www.mturk.com} for evaluating and testing \cite{chu2019meansum} their approach on YELP. We evaluate our model using these reference summaries as well.

\end{itemize}


In addition to the aforementioned datasets, the Amazon Product Review dataset with golden summaries is also used in \cite{bravzinskas2019unsupervised}. Specifically, Bra{\v{z}}inskas et al. curated a dataset with gold-standard summaries derived from the \textbf{Amazon Product Review} dataset, where 15 products from each of the Amazon review categories, i.e., Electronics, Clothing, Shoes and Jewelry, Home and Kitchen, Health and Personal Care, are sampled. On top of that, 8 reviews from each product are selected to serve as summaries. Three workers were assigned to each product with the task of reading the reviews and writing a summary text. These summaries are exclusively used for evaluation purposes. We assess our model, trained on our custom pre-processed Amazon dataset, using the corresponding summaries in our experiments.

In order to train our model on these datasets, we first need to pre-process them in accordance with the pipeline detailed in Section~\ref{subsec2}. This ensures that our model can be effectively trained using a self-supervised approach.


\subsection{Data Pre-processing}\label{subsec2}

\begin{figure}
    \centering
    \includegraphics[width=0.8\textwidth]{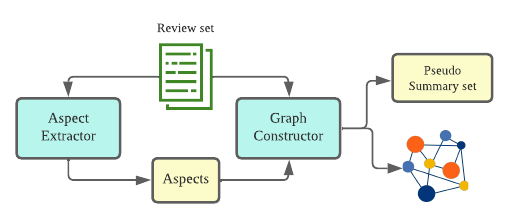}
    \caption{Data Pre-Processing} \label{fig3}
\end{figure}

Figure~\ref{fig3} demonstrates the process of data pre-processing. Basically, after extracting aspects from reviews, they are employed in conjunction with the review set to construct Knowledge Graphs and corresponding pseudo-summary sets. Three key stages are elaborated as follows. 

\begin{figure}
    \centering
    	\includegraphics[width=0.8\textwidth]{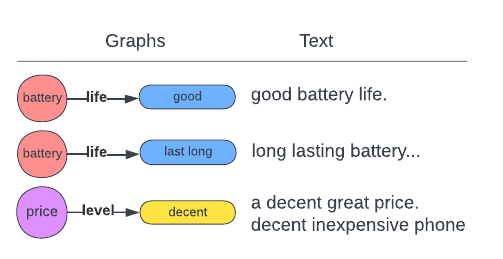}
    	\caption{The corresponding sentences are recorded and used as pseudo summaries for training and evaluation.} \label{fig4}
\end{figure}

\begin{itemize}
    \item \textbf{Data Cleaning Process.} Initially, the dataset undergoes a cleaning procedure, where entries with a review length of fewer than 100 characters are removed. While this may lead to the loss of some information from the overall review pool, the final performance of our model relies on the processed curated dataset. Any eliminated information is considered non-existent. Subsequently, we leverage product IDs to determine the number of products with more than five reviews, ensuring a sufficient quantity of information for each product. The chosen parameters, such as length and the number of reviews, may vary, and their impact will be assessed in future studies. The dataset is then partitioned into smaller subsets based on product categories. Furthermore, we eliminate the product ID column and merge the review headline and content for each entry, creating a single cohesive review text. We also remove undesired symbols, punctuation, URLs, HTML tags, and similar elements. The processed dataset serves as the foundation for pre-training a BERT model, which will generate sentence embeddings and attention matrices for both aspect extraction and Knowledge Graph construction. Additionally, the model will function as the text encoder in subsequent stages. 
    
    \item \textbf{Aspect Extraction and Knowledge Graph Construction.} The cleaned dataset is comprised solely of review texts. The subsequent steps involve extracting aspects from the dataset and constructing Knowledge Graphs for each product, incorporating aspects and related content. To extract aspects of each product, we utilize the pre-trained BERT model to generate embeddings for every sentence within the reviews. These sentence embeddings are then clustered, with each cluster's topic representing one aspect of the product. The number of clusters varies across products. For each product, we employ Spacy \footnote{https://spacy.io/} to extract noun chunks, and leverage the sentence attention embeddings to identify the central noun chunk of the sentence, subsequently merging similar chunks. The final chunk count determines the number of clusters. By utilizing the reviews within a cluster, we construct a Knowledge Graph for each product. Edges within these graphs are weighted by calculating the proportion of mentioned attributes, where Figure~\ref{fig1} further illustrates the weighted graphs. 

    \item \textbf{Sample Pairs Mapping.} Given that these datasets lack gold-standard summaries, it is not feasible to train the model in a supervised manner or evaluate the model using metrics like ROUGE \cite{lin2004rouge}. Therefore, our model is trained using a self-supervised approach. To create the dataset for self-supervised training, we extract one or multiple triplets from a single sentence during the Knowledge Graph construction process. We then randomly select $k$ aspects and merge the corresponding triplets into one graph. This graph is subsequently mapped with the pseudo summary, which consists of related sentences. The value of $k$ varies based on the different number of aspects for each product. Figure~\ref{fig4} displays the data samples in our opinion summarization dataset.
\end{itemize}

\subsection{Evaluation Metrics}\label{subsec3}

To evaluate our model, we use ROUGE \cite{lin2004rouge} to measure the lexical overlap between generated and reference summaries. Rouge\_1, Rouge\_2, and Rouge\_L are reported, representing uni-gram, bi-gram, and longest common subsequence overlaps, respectively.

In addition to ROUGE, we also assess our model using another metric called aspect coverage. This metric employs an aspect-based sentiment analysis (ABSA) task \cite{miao2020snippext} to predict the category and sentiment of extracted opinion phrases from summaries.

\subsection{Baselines}\label{subsec4}

\begin{itemize}
    
    \item \textbf{LexRank} \cite{erkan2004lexrank} is an unsupervised graph-based summarization method. It employs a ranking algorithm to determine node centrality. In LexRank, sentences are treated as nodes to form a graph with weighted edges calculated using tf-idf. In our work, following the settings from \cite{angelidis2021extractive}, we also use BERT \cite{devlin2018bert} and SentiNeuron \cite{radford2017learning} vectors to calculate the adjacency matrices.

    \item \textbf{Opinosis} \cite{ganesan2010opinosis} is a graph-based summarization framework that generates concise abstractive summaries of highly redundant opinions. It assumes no domain knowledge and leverages mostly the word order in the existing text.

    \item \textbf{Meansum} \cite{chu2019meansum} is an unsupervised neural model for multi-document summarization. It proposes an end-to-end architecture with an auto-encoder, where the mean of input review representations decodes into a reasonable summary review without relying on any review-specific features.

    \item \textbf{Copycat} \cite{bravzinskas2019unsupervised} is a summarization model based on the pointer-generator mechanism \cite{see2017get}. It follows the intuition of controlling the "amount of novelty" during summary generation. With this intuition, they define a hierarchical variational autoencoder model to produce summaries that reflect common opinions.

    \item \textbf{QT} \cite{angelidis2021extractive} enhances the ability to control the summarization process by leveraging the properties of quantized space to generate aspect-specific summaries.
\end{itemize}

\section{Experimental Results}\label{sec4}

In this section, we present the results of our model in comparison with other baselines. We evaluate our method alongside the baselines on both General and Aspect-adaptive Summarization to demonstrate its performance.

\subsection{Experiment 1: General Summarization}\label{subsec5}

We begin by presenting the results for General Summarization. In this experiment, we employ the full aspect set and do not control the weight, generating a summary that encompasses all aspects of a product. Aspects with low-weight labelled attributes are also included in the summary. A comparison of our model with other baselines on the SPACE dataset is shown in Table~\ref{tab1}. Considering the Aspect Coverage metric, our model outperforms the other baselines. However, when compared to human-written summaries, there remains a significant gap, indicating substantial room for improvement.

\renewcommand{\arraystretch}{1.5} 
\begin{table}[]
\caption{The results of the general summarization experiment on the SPACE dataset.}
\label{tab1}
\centering
\begin{tabular}{l|ccc|c}
\hline
SPACE(general)  & Rouge\_1       & Rouge\_2       & Rouge\_L       & Aspect Coverage(F1) \\ \hline
LexRank         & 29.85          & 5.87           & 17.56          & 0.520               \\
LexRank(SENTI)  & 30.56          & 4.75           & 17.19          & 0.520               \\
LexRank(BERT)   & 31.41          & 5.05           & 18.12          & 0.520               \\
Opinosis        & 28.76          & 4.57           & 15.96          & 0.570               \\
MeanSum         & 34.95          & 7.49           & 19.92          & 0.610               \\
Copycat         & 36.66          & 8.87           & 20.90          & 0.676               \\
QT              & 38.66          & 10.22          & 21.90          & 0.758               \\
AaKOS(ours)     & \textbf{39.42} & \textbf{11.06} & \textbf{23.48} & \textbf{0.772}      \\ \hline
Human Up. Bound & 49.80          & 18.80          & 29.19          & 0.845               \\ \hline
\end{tabular}
\end{table}

The comparison of our model with other baselines on the Amazon dataset is presented in Table~\ref{tab2}. With the exception of the Rouge\_L result, our model surpasses all baselines in performance. The performance comparison between our proposed model and several baseline models on the YELP dataset is detailed in Table~\ref{tab_yelp}. It is evident that our model demonstrates superior performance, outperforming all other baseline models under comparison.

\begin{table}[]
\caption{The results of the general summarization experiment on the Amazon dataset.}
\label{tab2}
\centering
\begin{tabular}{l|ccc|c}
\hline
Amazon(general) & Rouge\_1       & Rouge\_2      & Rouge\_L       & Aspect Coverage(F1) \\ \hline
LexRank(BERT)   & 31.47          & 5.07          & 16.81          & 0.663               \\
Opinosis        & 28.42          & 4.57          & 15.50          & 0.614               \\
MeanSum         & 29.20          & 4.70          & 18.15          & 0.710               \\
Copycat         & 31.97          & 5.81          & \textbf{20.16} & 0.731               \\
QT              & 34.04          & 7.03          & 18.08          & 0.739               \\ \hline
AaKOS(ours)     & \textbf{35.21} & \textbf{7.58} & 20.04          & \textbf{0.752}      \\ \hline
\end{tabular}
\end{table}

\begin{table}[]
\caption{The results of the general summarization experiment on the YELP dataset.}
\label{tab_yelp}
\centering
\begin{tabular}{l|ccc|c}
\hline
YELP(general) & Rouge\_1       & Rouge\_2      & Rouge\_L       & Aspect Coverage(F1) \\ \hline
LexRank(BERT)   & 26.46          & 3.00          & 14.36          & 0.601               \\
Opinosis        & 24.88          & 2.78          & 14.09          & 0.672               \\
MeanSum         & 28.46          & 3.66          & 15.57          & 0.713               \\
Copycat         & 29.47          & 5.26          & 18.09 & 0.728               \\
QT              & 28.40          & 3.97          & 15.27          & 0.722               \\ \hline
AaKOS(ours)     & \textbf{30.12} & \textbf{5.68} & \textbf{20.35}          & \textbf{0.736}      \\ \hline
\end{tabular}
\end{table}

Although the ROUGE results meet our expectations, the Aspect Coverage does not significantly outperform the other baselines as anticipated. In General Summarization, we apply all extracted aspects, and all relevant content in the Knowledge Graph is utilized, so every aspect should be included in the summary.

Two factors may limit the improvement of Aspect Coverage: \textbf{1)} aspect extraction: we employ clustering to identify aspects with a pre-defined number of clusters, which is likely to introduce bias. The impact of the number of clusters should also be evaluated; \textbf{2)} output length limitation: since we set an output length limit of 256, some information is inevitably omitted after reaching this constraint.

As the first factor is related to data pre-processing, which affects the entire training process and requires a considerable amount of time to complete, we only conduct an experiment on varying the output length to validate our assumption regarding the second factor. Results are displayed in Table~\ref{tab3}. As the table indicates, when the output length is increased from 256 to 512, our model exhibits a noticeable improvement, confirming that our assumption about the second factor is accurate. 

\begin{table}[]
\caption{The results of Aspect Coverage with output length 512.}
\label{tab3}
\centering
\begin{tabular}{l|c|c|c}
\hline
Models          & AC(SPACE) & AC(Amazon) & AC(YELP) \\ \hline
LexRank(BERT)         & 0.520      & 0.663     & 0.601               \\
Opinosis         & 0.570              & 0.614        & 0.672            \\
MeanSum         & 0.610                  & 0.710        & 0.713            \\
Copycat         & 0.676                  & 0.731       & 0.728             \\
QT              & 0.758                  & 0.739       & 0.722             \\
AaKOS(256)     & 0.772                  & 0.752        & 0.736           \\
AaKOS(512)     & \textbf{0.803}         & \textbf{0.824}     & \textbf{0.826}     \\ \hline
Human Up. Bound & 0.845                  & -            & -            \\ \hline
\end{tabular}
\end{table}

\subsection{Experiment 2: Aspect-adaptive Summarization}\label{subsec6}

In this experiment, we demonstrate the ability of our model to adapt to various aspects on the SPACE dataset. Table~\ref{tab4} compares six specific aspects individually based on Rouge\_L, and average results for Rouge\_1, Rouge\_2, and Rouge\_L are also shown. To ensure that the outputs of other baselines contain only the content of the particular aspect, we use the aspect to filter out the relevant sentences from our pre-processed SPACE dataset, as described in Section~\ref{subsec1}. These filtered contents are then used as input for all baseline models. Since our model requires different inputs, we utilize the corresponding graphs and aspects as inputs.

The result for the ``Rooms" aspect falls below the QT's result, and for ``Food", it only slightly surpasses the best baseline. On other aspects and average results, our model outperforms all baselines, exhibiting significant improvement, particularly on ``Cleanliness". However, a considerable gap between our model and human-written results still exists.

\begin{table}[]
\caption{The results of aspect-adaptive summarization experiment on SPACE dataset.}
\label{tab4}
\centering
\begin{tabular}{l|cccccc|ccc}
\hline
\multicolumn{1}{c|}{\multirow{2}{*}{SPACE(ASP)}} & \multicolumn{6}{c|}{Rouge\_L}                                                                       & R\_1           & R\_2          & R\_L           \\
\multicolumn{1}{c|}{}                            & Building       & Cleanliness    & Food           & Location       & Rooms          & Service        & \multicolumn{3}{c}{\textbf{Average}}            \\ \hline
LexRank(ASP)                                     & 11.56          & 17.65          & 11.73          & 16.64          & 15.23          & 18.65          & 21.46          & 3.26          & 15.24          \\
MeanSum(ASP)                                     & 15.67          & 14.03          & 13.25          & 19.03          & 15.67          & 18.14          & 21.87          & 4.48          & 15.97          \\
Copycat(ASP)                                     & 13.28          & 22.64          & 15.25          & 21.59          & 16.80          & 24.62          & 24.35          & 7.16          & 19.03          \\
QT(ASP)                                          & 15.31          & 21.38          & 16.03          & 22.16          & \textbf{19.83} & 24.38          & 26.18          & 7.85          & 19.85          \\ \hline
AaKOS(ASP)                                      & \textbf{17.12} & \textbf{25.06} & \textbf{16.29} & \textbf{24.02} & 19.58          & \textbf{26.13} & \textbf{28.67} & \textbf{8.65} & \textbf{21.37} \\ \hline
Human                                            & 40.33          & 38.76          & 33.63          & 35.23          & 29.25          & 30.31          & 44.86          & 18.45         & 34.58          \\ \hline
\end{tabular}
\end{table}

\begin{table}[]
\caption{The general summaries from all models about ``Hotel Erwin".}
\label{tab5}
\centering
\begin{tabular}{l|l}
\hline
\textbf{Models} & \textbf{General Summaries for ``Hotel Erwin"}                                                                          \\ \hline
Human           & \begin{tabular}[c]{@{}l@{}}All staff members were friendly, accommodating, and helpful. The hotel and room \\ were very clean. The room had modern charm and was nicely remodeled. The beds \\ are extremely comfortable. The rooms are quite with wonderful beach views. The \\ food at Hash, the restaurant in lobby, was fabulous. The location is great, \\ very close to the beach. It’s a longish walk to Santa Monica. The price is \\ very affordable.\end{tabular} \\ \hline
MeanSum         & \begin{tabular}[c]{@{}l@{}}It was a great stay! The food at the hotel is great for the price. I can’t \\ believe the noise from the street is very loud and the traffic is not so great, \\ but that is not a problem. The restaurant was great and the food is excellent.\end{tabular}                                                                                                                                                                                     \\ \hline
Copycat         & \begin{tabular}[c]{@{}l@{}}This hotel is in a great location, just off the beach. The staff was very \\ friendly and helpful. We had a room with a view of the beach and ocean. The only \\ problem was that our room was on the 4th floor with a view of the ocean. If you \\ are looking for a nice place to sleep then this is the place for you.\end{tabular}                                                                                                           \\ \hline
QT              & \begin{tabular}[c]{@{}l@{}}Great hotel. We liked our room with an ocean view. The staff were friendly and \\ helpful. There was no balcony. The location is perfect. Our room was very quiet. \\ I would definitely stay here again. You’re one block from the beach. So it must \\ be good! Filthy hallways. Unvacuumed room. Pricy, but well worth it.\end{tabular}                                                                                                       \\ \hline
AaKOS           & \begin{tabular}[c]{@{}l@{}}The Hotel Erwin is an excellent choice for its great and convenient location right \\ on Venice Beach. Rooms were quiet, clean, comfortable and with great ocean views, \\ even though some claimed the noise levels. The rooftop bar provides a great view \\ of the ocean and the experience of watching the sunset. The service of Hotel Erwin \\ is friendly and accommodating. Food and drinks are delicious at fare price.\end{tabular}    \\ \hline
 
\end{tabular}
\end{table}

\begin{table}[]
\caption{The aspect-adaptive summaries from AakOS about ``Hotel Erwin".}
\label{tab6}
\centering
\begin{tabular}{l|l}
\hline
\textbf{Aspects}      & \textbf{Aspect-adaptive Summaries for Hotel Erwin by AaKOS}                                             \\ \hline
Room and Service      & \begin{tabular}[c]{@{}l@{}}The rooms of Hotel Erwin were quiet, clean, comfortable and with great \\ ocean views. They have a modern and spacious design with comfortable \\ beds. However, bathrooms with sliding glass doors may have privacy \\ concerns. The hotel offers great, friendly and accommodating service with \\ helpful staff.\end{tabular} \\ \hline
Building and Location & \begin{tabular}[c]{@{}l@{}}The Hotel Erwin has a very great and convenient location. It locates right \\ on Venice Beach. The building is decorated newly and very stylish. The \\ rooftop bar provides a great view of the ocean and the experience of watching \\ the sunset.\end{tabular}                                                                   \\ \hline
Cleanliness and Food  & \begin{tabular}[c]{@{}l@{}}The hotel is clean. The rooms are clean, comfortable and well-maintained. \\ The hotel and the restaurant off the lobby provide great food and drinks \\ are delicious with fare price including breakfast, and room service.\end{tabular}                                                                                         \\ \hline
\end{tabular}
\end{table}

\subsection{Case Study} 
 


In this section, we present a case study by delving into a detailed comparison of the general and aspect-adaptive summarization capabilities of various models using the reviews of ``Hotel Erwin".

In Table~\ref{tab5}, we demonstrate the general summaries produced by several models, including the human-written summary and the one generated by the proposed AaKOS model. Upon examining the table, it is evident that the summaries provided by the AaKOS model align more closely with the human-written summary in terms of detail and comprehensiveness.

One of the noticeable differences is the perspective from which the summaries are written. Except for the human-written summary and that generated by AaKOS, all other models use a first-person perspective, as indicated by the frequent use of pronouns like ``I" and ``we". However, a summary should ideally represent the collective opinion of numerous reviews rather than an individual perspective. This highlights one of the strengths of the AaKOS model, i.e., its text-to-graph transformation process effectively filters out irrelevant nouns such as personal pronouns, retaining only the ones relevant to the aspects being summarized.

Moving to Table~\ref{tab6}, we can observe the aspect-adaptive summaries generated by AaKOS for ``Hotel Erwin". These aspect-oriented summaries provide detailed insights into specific features of the hotel, such as ``Room and Service", ``Building and Location", and ``Cleanliness and Food".

A noteworthy observation here is the depth and precision of the aspect-specific summaries produced by AaKOS. It brings out the nuances of different aspects like the privacy concerns related to bathroom designs under ``Room and Service" or the enjoyable experience of watching the sunset from the rooftop bar under ``Building and Location". Such focused summaries would be particularly beneficial for potential customers looking for information on specific aspects of the hotel.

This case study shows the effectiveness of the AaKOS model in generating both general and aspect-specific summaries. It effectively condenses broad opinions and provides detailed, relevant summaries. Moreover, its ability to create aspect-oriented summaries proves invaluable in providing potential customers with targeted information, ultimately aiding their decision-making process.

\section{Conclusions and Future Works}\label{sec13}



In this research work, we proposed a novel approach to conduct aspect-adaptive and knowledge-based opinion summarization, through the development of the Aspect-adaptive Knowledge-based Opinion Summarization (AaKOS) model. The AaKOS model is self-supervised, training on accurately matched pairs of aspect graphs and pseudo summaries. It proves effective in capturing diverse aspects from reviews and tailoring the summaries to align with the specific requirements of users.

To achieve this, reviews are first transformed into knowledge graphs, providing a structured representation of the information contained in the review. If users specify certain aspects they're interested in, the model retrieves the corresponding sub-graphs and leverages these to produce summaries that directly address the desired aspects. The model also introduces a weight controller that aids in accounting for varying sentiment trends, lending a dynamic dimension to the summarization.

Extensive experiments have been conducted to evaluate the performance of AaKOS model under both general text summarization and aspect-adaptive summarization tasks, where three real-world datasets are adopted, i.e., Amazon product review, SPACE, and YELP. The experimental results explicitly demonstrate its superiority over other baseline models in crafting comprehensive general summaries. These summaries span all aspects of reviews. Additionally, AaKOS model excels in generating targeted, aspect-specific custom summaries.

In the future, we will focus on further refining the AaKOS model. We aim to enhance the aspect extraction process and the creation of high-quality knowledge graphs. Improved aspect extraction would enhance the amount and relevance of information that can be retrieved from the dataset, while superior knowledge graphs would provide more accurate content to inform the summarization process.

Moreover, we plan to broaden the applications of our model by testing its capabilities in other summarization tasks. In particular, multi-document summarization, where information is condensed from multiple documents, presents a promising avenue for future exploration.

\section*{Data availability} 
The datasets analysed during the current study are publicly available at the following links:
\begin{itemize}
\item \textbf{Amazon Product Review dataset} is available at Amazon with the link: https://s3.amazonaws.com/amazon-reviews-pds/tsv/index.txt.

\item \textbf{SPACE} dataset is available and open at https://drive.google.com/open?id=1C6SaRQkas2B-9MolbwZbl0fuLgqdSKDT\&authuser=0.

\item \textbf{YELP} dataset is collected and published by Yelp as Yelp Open Dataset at https://www.yelp.com/dataset.

\end{itemize}

\newpage
\bibliography{ref}

\end{document}